\documentclass[conference]{IEEEtran}
\IEEEoverridecommandlockouts
\usepackage[switch]{lineno}
\usepackage{cite}
\usepackage{amsmath,amssymb,amsfonts}
\usepackage[noend]{algpseudocode}
\usepackage{listings}
\usepackage{graphicx}
\usepackage{textcomp}
\usepackage{xcolor}
\usepackage{lineno,hyperref}
\usepackage{subcaption}
\usepackage{array}
\usepackage{makecell}
\usepackage{multirow}
\usepackage{hyperref}
\usepackage[normalem]{ulem}
\usepackage{algorithm}
\usepackage{algpseudocode}
\usepackage{verbatim} 
\usepackage{tikz}
\usepackage{eso-pic}

\newcommand\copyrighttext{%
  \footnotesize \textcopyright 2021 IEEE. Personal use of this material is permitted.
  Permission from IEEE must be obtained for all other uses, in any current or future
  media, including reprinting/republishing this material for advertising or promotional purposes, creating new collective works, for resale or redistribution to servers or lists, or reuse of any copyrighted component of this work in other works. DOI: not yet}
\newcommand\copyrightnotice{%
\begin{tikzpicture}[remember picture,overlay]
\node[anchor=south,yshift=10pt] at (current page.south) {\fbox{\parbox{\dimexpr\textwidth-\fboxsep-\fboxrule\relax}{\copyrighttext}}};
\end{tikzpicture}%
}

\DeclareMathOperator*{\argmin}{argmin}

\def\BibTeX{{\rm B\kern-.05em{\sc i\kern-.025em b}\kern-.08em
    T\kern-.1667em\lower.7ex\hbox{E}\kern-.125emX}}
\begin{document}

\title{Multi-Objective Evolutionary Design of Composite Data-Driven Models
}

\makeatletter
\newcommand{\linebreakand}{%
  \end{@IEEEauthorhalign}
  \hfill\mbox{}\par
  \mbox{}\hfill\begin{@IEEEauthorhalign}
}
\makeatother

\author{\IEEEauthorblockN{Iana S. Polonskaia}
\IEEEauthorblockA{\textit{Natural Systems Simulation Lab} \\
\textit{ITMO University}\\
Saint-Petersburg, Russia \\
ispolonskaia@itmo.ru}
\and
\IEEEauthorblockN{Nikolay O. Nikitin}
\IEEEauthorblockA{\textit{Natural Systems Simulation Lab} \\
\textit{ITMO University}\\
Saint-Petersburg, Russia \\
nnikitin@itmo.ru}
\and\IEEEauthorblockN{Ilia Revin}
\IEEEauthorblockA{\textit{Natural Systems Simulation Lab} \\
\textit{ITMO University}\\
Saint-Petersburg, Russia}
\and
 \linebreakand 
 \IEEEauthorblockN{Pavel Vychuzhanin}
\IEEEauthorblockA{\textit{Natural Systems Simulation Lab} \\
\textit{ITMO University}\\
Saint-Petersburg, Russia}
 \and
\IEEEauthorblockN{Anna V. Kalyuzhnaya}
\IEEEauthorblockA{\textit{Natural Systems Simulation Lab} \\
\textit{ITMO University}\\
Saint-Petersburg, Russia}
}

\maketitle
\thispagestyle{plain}
\copyrightnotice
\pagestyle{plain}

\begin{abstract}
In this paper, a multi-objective approach for the design of composite data-driven mathematical models is proposed. It allows automating the identification of graph-based heterogeneous pipelines that consist of different blocks: machine learning models, data preprocessing blocks, etc. The implemented approach is based on a parameter-free genetic algorithm (GA) for model design called GPComp@Free. It is developed to be part of automated machine learning solutions and to increase the efficiency of the modeling pipeline automation. A set of experiments was conducted to verify the correctness and efficiency of the proposed approach and substantiate the selected solutions. The experimental results confirm that a multi-objective approach to the model design allows us to achieve better diversity and quality of obtained models. The implemented approach is available as a part of the open-source AutoML framework FEDOT.
\end{abstract}


\section{Introduction}
\label{sec_intro}

The design of data-driven mathematical models is an actual research direction in modern data science \cite{caldwell2013mathematical}. The internal structure of the model depends on the type of the learning algorithm, so complex data-driven models can consist of several semi-independent blocks - this approach is usually referred to as ensembling \cite{pavlyshenko2018using}. There are several techniques to build complex models: for example, blending allows creating single-level ensembles of machine learning (ML) models, and stacking allows creating multi-level ones. Other approaches are based on the representation of a model structure (or even the whole modeling pipeline) as a directed acyclic graph (DAG).

The selection of the most suitable model design is the core problem in automated machine learning (AutoML) \cite{feurer2018towards}. Until recently, successful machine learning pipelines have been designing manually by experts. However, this method requires performing many routine operations and repeating experiments which is generally a computationally expensive procedure that does not always lead to success. 

The aim of the model design search is to achieve the appropriate values of different criteria (e.g. quality of the modeling) and satisfy the constraints (e.g. time-related ones). In many cases, it is impossible to maximize all criteria simultaneously, which leads to a multi-objective optimization problem. The usage of  multi-objective approaches in AutoML solutions is a quite promising direction that can lead to better suitability of the obtained ML pipelines and even makes it possible to co-design the models and infrastructure \cite{kalyuzhnaya2021towards}.

In this paper, a flexible and effective multi-objective evolutionary approach (EA) to the composite model design is proposed and implemented as a part of the automated modeling framework FEDOT. We modified the standard evolutionary scheme with a self-configuration heuristic for the hyper-parameters (such as population size, crossover and mutation probabilities, maximal model structure graph size). The performed experiments demonstrate that the implemented approach named GPComp@Free provides an ability to design effective data-driven models from scratch.


\section{Related work}
\label{sec_related}

There is a variety of approaches that can be used to identify the optimal design of a data-driven model. For instance, AutoML solutions can be based on random search \cite{ledell2020h2o}, Bayesian optimization \cite{feurer2019auto}, reinforcement learning (RL) \cite{heffetz2020deepline}, Monte Carlo tree search \cite{rakotoarison2019automated}, gradient-based approaches \cite{liu2018darts}. However, most of them are less flexible than EAs to the model design (implemented e.g. in \cite{qi2020graph}). Their conceptual advantage is open-endedness  \cite{packard2019open}, which allows "growing" of the model with the most suitable structure instead of the direct search of appropriate solutions in some restricted range of structural parameters. 


Furthermore, evolutionary algorithms (EA) are much more flexible due to their simplicity and clarity and can be easily extended with new evolutionary operations. Thus, in \cite{real2019regularized} the authors demonstrate a tournament selection modification with memory (which contains the best genotypes) based on the association of genotypes with their age, and displacement of the tournament selection to choose the ``younger'' genotypes. This modification allows us to obtain similar results in convolution neural network architecture search problems faster than when using well-established reinforcement learning (RL) algorithms. Another case when the EA competes with RL approaches is presented in \cite{salimans2017evolution}. Authors demonstrate the advantages of evolution strategies in comparison with popular RL techniques (such as Q-learning and policy gradients) on test problems from the OpenAI Gym library. 


Multi-objective optimization is implemented in several existing AutoML solutions, e.g. TPOT \cite{ olson2016tpot} and autoxgboostmc \cite{pfisterer2019multi}. However, the diversity of the obtained solutions is quite limited in these tools (because the custom graph-based model structure is not supported). Also, multi-objective optimization is widely used as a part of Neural Architecture Search approaches (NAS). For example, in \cite{lu2019nsga} the second objective function is usually based on structural complexity. Besides the model structure design, the multi-objective hyperparameters tuning can be applied to ML \cite{binder2020multi} and domain-specific models. 

Various criteria can be involved in the multi-objective model design. Besides the basic characteristics like quality and complexity, there are additional objectives that can be used, e.g. fairness, interpretability, robustness, and the sparsity of models (as described in \cite{pfisterer2019multi}). However, in \cite{pfisterer2019multi} multi-objective optimization is applied to a single machine learning model and graph-based structures are not discussed.

It can be concluded that the existing methods in this field are quite specialised, which leads us to the aim of developing a more universal and effective multi-objective approach that can be used to create an effective design of the various data-driven models with graph-based structure.


\section{Problem statement}
\label{sec_problem}

There is a lot of different modeling tasks that can be solved using machine learning: classification, clustering, regression, time series forecasting, ranking. However, all of them can be formulated as results of evaluation of the model $M$.

In a common case, the structure of the model $M$ can be represented as a graph that consists of several atomic models. An atomic model is a machine learning model that has a logically indivisible structure in terms of a particular problem. In fact, an atomic model can have an arbitrary internal structure (e.g. an ensemble obtained by gradient boosting). It is a quite flexible representation that allows us to describe different types of models beyond machine learning: hybrid \cite{kalyuzhnaya2020automatic}, equation-based \cite{hvatov2020}, etc. In this case, $M$ can be formalized as a composite model - a data-driven model with an explicit internal structure, where several atomic models can be identified.

Since the data flow is directed from the input $X$ of the model to the output $X$ with the general statement of numerical simulation problem, the internal structure of the model $M$ can be represented as a directed acyclic graph (DAG) ${M}^{G}$ and defined as follows:

\begin{equation}
\begin{aligned}[t]
	Y=\mathcal{H}(M|X), {{M}^{G}}=\left\langle {{V}_{i}},{{E}_{j}} \right\rangle =\left\langle {{A}_{i}},{{\{{{H}_{{{A}_{i}}}}\}}_{k}},{{E}_{j}} \right\rangle,
\end{aligned}
\end{equation}	
where $\mathcal{H}$ is a modelling operator for the model $M$ with input data $X$ and $Y$ as model results. For ML models, the data set $X$ is usually split into subsamples $X_{train}$ and $X_{test}$. In the graph model ${M}^{G}$, the vertices $V$ have complex structure and can be represented as tuple $\left\langle {{A}_{i}},{{\{{{H}_{{{A}_{i}}}}\}}_{k}}\right\rangle$ with different data-driven models ${A}_{i}$ and its hyper-parameters ${{\{{{H}_{{{A}_{i}}}}\}}_{k}}$. Directed edges $E$ represent the data flow from income vertices (the model's input) to outcome vertices (the model's output).

The application of the evolutionary optimization approaches leads to the following interpretation: ${M}^{G}$ structure can be considered as the genotype and phenotype is expressed through integral characteristics of realization ${M}^{G}$. In the frame of this paper, the integral characteristics fill the objective vector function $F$.  

As can be seen from Sec.~\ref{sec_related}, there are different objective functions that can be used in the design of the composite model: modeling quality for a specific data set, structural complexity, computation time, interpretability, robustness, number of features required to build the model (related to the sparsity). Also, specific infrastructure-related objectives can be involved as part of the co-design and scheduling tasks \cite{chirkin2017execution}. In this paper, we decided to consider the quality, structural complexity, and computational performance objectives.

In this case, the formulation of the optimization task can be defined as follows:
\begin{eqnarray}
{M}^{G}_{opt} = \argmin_{{M}^{G}}{F({A}_{i}},{{\{{{H}_{{{A}_{i}}}}\}}_{k},{E}_{j})}, \\
\label{eq_finness}
F({M}^{G}) = (Q({M}^{G}), S({M}^{G}), P({M}^{G})), \nonumber\\ {F}_{hyp} = HV(F({M}^{G})),
\end{eqnarray}

where $F$ is a vector objective function, $Q$ is a quality-based criterion for the prediction obtained by the model ${M}^{G}$, $S$ is the structural complexity criteria (tightly connected with the interpretability of the model), $P$ is the computational performance criterion of the model. Here we miss arguments $X$ and $Y$ for compactness of formulas and because these conditions remain constant while optimization proceeds. Additional measure ${F}_{hyp}$ that helps to join components of vector objective function is Pareto hypervolume operator $HV$. 

There are different interpretations of the quality of the multi-objective design that can be used. In the paper, we used the following set of criteria: the root-mean-seared-error (RMSE) as $Q$ criterion for the regression problems, the negative area under the ROC curve ((-1)*(ROC AUC)) as $Q$ criterion for the classification problems, fitting time as the computational complexity criterion $P$, graph size $G_s$ as a structural complexity criterion $S$, and hypervolume \cite{cao2015using} as a quality measure for the Pareto frontiers.

The possible alternative to the 'real' multi-objective approach is the application of penalty-based optimization algorithm. In this case, the Eq.~\ref{eq_finness} can be formulated as follows:

\begin{equation}
\label{eq_penalty}
\begin{aligned}[t]
	F({M}^{G}) = Q({M}^{G}) - S({M}^{G})w_{1} - P({M}^{G})w_{2},
\end{aligned}
\end{equation}	
where $w_{1}$ and $w_{1}$ are weights for the complexity penalties. However, this implementation can affect the diversity during the optimization and lead to insufficient efficiency of the model design.

The described statement of the composite model problem is summarized in Figure~\ref{fig_concept}.

\begin{figure*}[t!]
\centerline{\includegraphics[width=16cm]{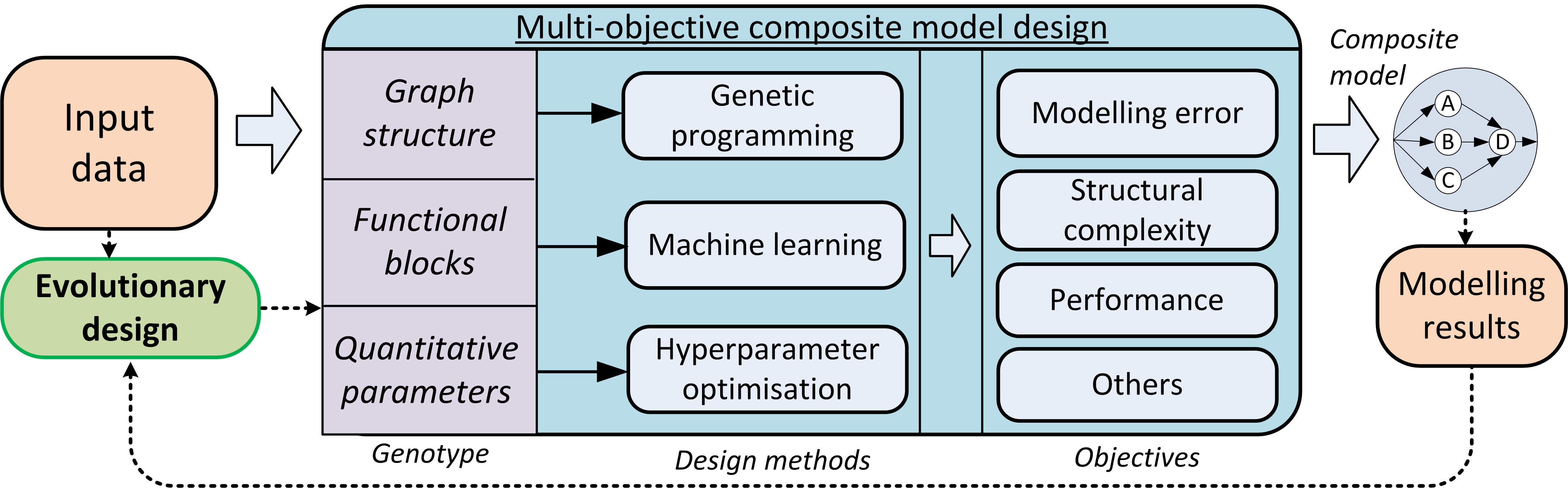}}
\caption{The concept of the multi-objective evolutionary design of the composite machine learning model. The internal structure of the model, design approaches, possible objectives and their interactions are presented as separate blocks to highlight the main issues of the optimization.}
\label{fig_concept}
\end{figure*}

In this paper, the following hypotheses can be stated:

\subsubsection*{Hypothesis 1}\label{hypmo} The use of multi-objective optimization as a part of composite models identification methods allow us to obtain the models with better quality than single-objective approaches due to the more effective preservation of the diversity.

\subsubsection*{Hypothesis 2}\label{hypsel} The selection of the appropriate genetic operators makes it possible to increase the quality of the obtained solutions for different cases.

\subsubsection*{Hypothesis 3}\label{hypself} The hyper-parameters of the multi-objective evolutionary algorithm can be self-configured during the evolution using heuristic techniques without the computationally expensive meta-optimization methods.

\section{Multi-objective design of composite models}
\label{sec_design}

\subsection{Evolutionary optimization of the models structure}

There are many extensively used conventional optimization techniques, but evolutionary algorithms deserve particular attention due to their range of advantages. They have a lower probability of getting stuck in a local optima than most optimization algorithms because they perform a global search in the solution space. In addition, such algorithms can be easily parallelized. This property is especially important for applied engineering problems since the time for finding the solution is usually restricted. Moreover, evolutionary algorithms are easily interpreted for almost any problem, therefore the user of an evolutionary-based framework always has an opportunity to adapt it to another optimization problem (e.g. to the problem of the neural architecture search). 

The proposed implementation of composite models structure optimizer is based on genetic programming (GP) algorithms \cite{winkler2005new}. Unlike many other algorithms, genetic programming allows us to use a high-level problem statement. It finds a composite model (or several candidate models in the multi-objective case) for a certain machine learning problem (classification, regression and time series forecasting problems are currently supported) from a pre-defined set of available models.  


The algorithm initializes the population of chains with random structure at the first step. Next, a few models in the population are selected as the parents using a preferred selection method. Each new model (offspring) is constructed from two parents by the application of a crossover operator, and some of them are impacted by a transformation called a mutation. Each operator has a performing probability which can be static or dynamic depending on the chosen evolutionary scheme. Once the child model structure is produced, it is then fitted with a training sample, evaluated with a test sample, and added to the new population. These manipulations are performed during several generations until the termination criteria are satisfied. 

 In experiments we used subtree crossover which replaces random subtree in one parent to random subtree from another and one-point crossover which performs in the same way, but subtrees is selected from common structural parts between two trees. Three mutation's options: simple, subtree and  reduce (first option changes models in random nodes, second changes random subtree to randomly generated and last removes random subtree). In \cite{nikitin2020structural} we described in detail the composite model representation in the genetic programming algorithm GPComp and how the custom evolutionary operators are used. Thus, in this paper, we focus on the aspects of multi-objective evolutionary optimization technique, describe implemented evolutionary schemes, and demonstrate its effectiveness on certain data sets for machine learning algorithms benchmarking.


The composite model optimization algorithm GPComp includes two of the most common evolutionary schemes: steady-state (also named ($\mu+\lambda$)) and generational. In the first scheme, the new population is generated by using a selection operator which is applied to the union of the offspring and the previous population. In the generational scheme the offspring completely replaces the parent population. Our previous experiments (also in \cite{nikitin2020structural}) demonstrate the clear domination of steady-state scheme over the generational. Thus, we didn't apply the generational scheme in this paper. 

\subsection{GPComp@Free: algorithm for parameters-less multi-objective evolution of composite models}

Although the AutoML concept focuses on reducing (or excluding altogether) human participation in the optimization process, the majority of the state-of-the-art tools raise an additional meta-optimization problem related to the identification of their most-effective hyperparameters. It is well-known that the EA has been solving many real-world optimization problems successfully, but their performance can depend considerably on the particular hyperparameters values used. Hyperparameters self-configuration procedures allow us to achieve complete automation. Our parameter-free multi-objective genetic scheme (GPComp@Free) based on an adaptive EA (described in \cite{evans2020adaptive}) allows configuring most of EA hyperparameters through the algorithm execution unlike aforementioned GPComp which suppous hand-tuning of hyperparameters. The proposed algorithm uses a steady-state evolutionary scheme, but $\mu$ (population size) changes during evolution like the Fibonacci sequence and $\lambda$ always equals to the previous item of the sequence with respect to $\mu$. Crossover and mutation rates ($c_{rate}$, $m_{rate}$) are also changed depending on the diversity of the population. The scheme of this approach is presented in Figure~\ref{fig_param_free_scheme}. The approach assumes two optimization objectives are used: solution quality (main goal) and chain complexity. We used two conditions for the hyperparameters changes: if none of the objective function values is improved then $\mu$ is increased to the next number in the sequence and, if both objectives (quality and complexity) are improved in the offspring population (with respect to their parents) then $\mu$ is decreased to the previous number in the sequence and crossover and mutation rates change according to the rule described in the AdaptedEvoParams procedure (Alg.~\ref{alg_gp}).

\begin{figure}[h]
\centerline{\includegraphics[width=9cm]{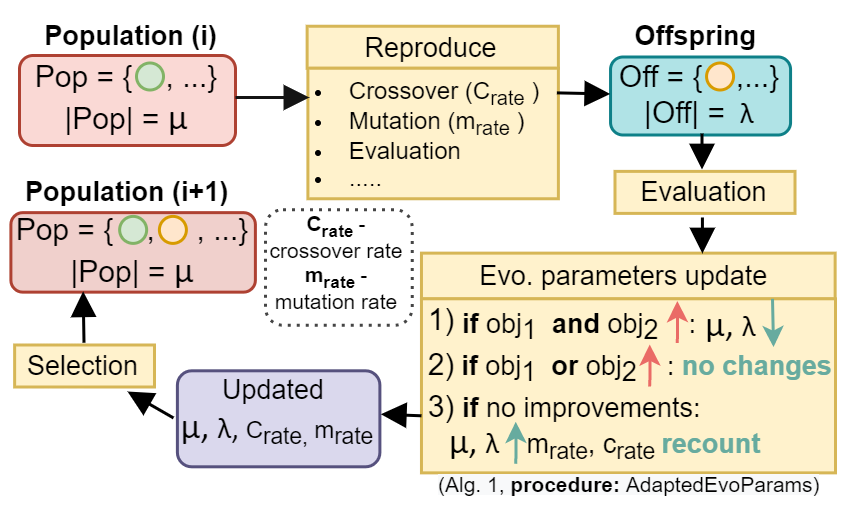}}
\caption{The description of one iteration of the parameter-free evolutionary algorithm. Population size ($\mu$), offspring size ($\lambda$), crossover and mutation rates ($c_{rate}$, $m_{rate}$) adapt during the algorithm execution.}
\label{fig_param_free_scheme}
\end{figure}

Also, we implemented the heuristics adaptation of chains' depth which allows avoiding over-complicated individuals. The algorithm with depth adaptation starts from low-depth chains and if there are no possibilities to improve objectives (stagnation during several generations) then depth is increased (see Alg.~\ref{alg_gp}, DepthAdaptation procedure). The details of the parameter-free optimizer implementation (further in the paper it will be called GPComp@Free) are described in Alg.~\ref{alg_gp}. The version without depth adaptation will be called "GPComp@Free with fixed depth" in the paper. Both algorithms are implemented using DEAP \cite{de2014deap}.

\begin{algorithm*}
\caption{GPComp@Free (parameter-free implementation of the multi-objective GP algorithm for the model design)}\label{gpcompfree}

\begin{algorithmic}[1]

\Procedure{ParameterFreeComposer}{models, data, objectiveFunctions, timeConstraint}  
    \State \underline{Input:} $ objectiveFunctions$ = $\left\{qualityObjective,  complexityObjective \right\}$ 
    \State \underline{Output:} array with nondominated models (Pareto frontier)
    \State $stagnationThreshold \gets 5$ \Comment {Allowed number of populations without improvement before max depth update}
    \State $maxDepth \gets 2$ \Comment{Inital values of chain depth}
    \State $SequenceFunction \gets FibonacchiIterator$ \Comment{Heuristic for the population size}
    \State $popSize \gets SequenceFunction.StateByIndex(index = 2)$ \Comment{Default size equals to second sequence item} 
    \State $crossRate, mutRate \gets 0.5, 0.5$
    \State $pop\gets \Call{InitPopulation}{models, popSize} $
    \State $pop\gets \Call{Evaluate}{pop, objectiveFunctions, data}$
    \State $currentStd, maxStd  \gets \Call{QualityStdCalculation}{pop}$  \Comment{Standard deviation for prediction quality objective}
    \While{$time() < timeConstraint$}
        \State $offspringSize \gets SequenceFunction.PreviousState(popSize)$
        \State $pareto, selectedParents \gets \Call{Updatepareto}{pop}, \Call{MultiObjSelection}{pop, offspringSize}$
        \State $offspring \gets \Call{Reproduct}{selectedParents, offspringSize, crossRate, mutRate, maxDepth, models}$
        \State $pop \gets \Call{Evaluate}{offspring, objectiveFunctions, data}$
        \State $stagnationCnt, maxDepth \gets \Call{DepthAdaptation}{stagnationCnt, pareto, stagnationThreshold, maxDepth}$
        \State $mutRate, crossRate, popSize \gets \Call{AdaptedEvoParams}{offspring, pareto, popSize, mutRate, crossRate}$
        \State $pop \gets \Call{MultiObjSelection}{pop\cup offspring, popSize}$
        \State $currentStd \gets \Call{QualityStdCalculation}{pop}$
        \State $maxStd \gets \Call{UpdateMaxStd}{currentStd}$
    \EndWhile\label{constraintwhile}
    \State \textbf{return} $pareto$ 
\EndProcedure \label{optimiser}
\Procedure {DepthAdaptation}{stagnationCnt, pareto, stagnationThreshold, maxDepth}
    \State $stagnationCnt \gets \Call{UpdatestagnationCnt}{pareto}$ \Comment{Number of generations without Pareto frontier's changes}
    \If{$stagnationCnt == stagnationThreshold$}
        \State $maxDepth \gets maxDepth + 1$ \Comment{Increase the maximum allowed depth of the composite model structure graph}
    \EndIf
    \State \textbf{return} $stagnationCnt, maxDepth$
\EndProcedure
\Procedure {AdaptedEvoParams}{offspring, pareto, popSize, mutRate, crossRate}
    \If{\Call{NoObjectivesImprovements}{offspring, pareto}}
        \State $popSize \gets SequenceFunction.NextState(popSize)$
        \State $mutRate, crossRate \gets 1-(currentStd/maxStd), currentStd/maxStd$
    \ElsIf {\Call{AreBothObjectivesImproved}{offspring, pareto}}
        \State $popSize \gets SequenceFunction.PreviousState(popSize)$ 
    \EndIf
    \State \textbf{return} $mutRate, crossRate, popSize$
\label{evo_params_adapt}
\EndProcedure
\end{algorithmic}
\label{alg_gp}
\end{algorithm*}

The described approach is implemented as a part of the automated modeling framework FEDOT \cite{kalyuzhnaya2020automatic}. It supports both single- and multi-objective optimization of the ML pipelines that includes the composite models. The main feature of the framework is the complex management of interactions between various computing elements of the pipelines. The framework can be used to automate the creation of mathematical models for various problems, different types of data, and models. It is available under the open-source license in \url{https://github.com/nccr-itmo/FEDOT}.


\section{Experimental studies}
\label{sec_exp}

\subsection{Setup of experiments}

To analyze the correctness and effectiveness of the proposed approach, a set of experiments was conducted using several data sets from the Penn Machine Learning Benchmarks repository \cite{olson2017pmlb}. These data sets cover a broad range of applications, and combinations of categorical, ordinal, and continuous features. There are no missing values in these data sets. Selected data sets were split on training and test sets in the ratio of 70/30.

Key features of the selected data sets are shown in \ref{tab_dataset_features}. Feature 'Imbalance' show   a value of imbalance metric, where zero means that the data set is perfectly balanced and the higher the value, the more imbalanced the data set.



\begin{table}
\centering
\caption{Key features of the selected data sets}
\label{tab_dataset_features}
\begin{tabular}{|c|c|c|c|c|} 
\hline
Data set      & $N_{samples}$ & $N_{features}$ & Task type   & Imbalance  \\ 
\hline
Churn        & 5000          & 20             & binary clf. & 0.51       \\ 
\hline
Dis          & 3772          & 29             & binary clf. & 0.93       \\ 
\hline
Hill\_Valley & 1212          & 100            & binary clf. & 9.81       \\ 
\hline
Elusage      & 55            & 2              & regr.       & -          \\
\hline
\end{tabular}
\end{table}

There were three complex experiments prepared:

\subsubsection*{Experiment 1} Comparison of the various model design approaches. The single-objective genetic algorithm (GPComp), a single-objective genetic algorithm with penalty-based fitness (described in Eq.~\ref{eq_penalty}), and the multi-objective algorithm GPComp@Free were used. The aim of the experiment is to ensure that the multi-objective approach can find models with equal or better prediction quality than the single-objective approach (see \hyperref[hypmo]{Hypothesis 1}).
\subsubsection*{Experiment 2} Comparison of different implementations of the GPComp@Free algorithm. Two variants of the algorithm are based on different strategies inspired by well-known algorithms SPEA2 \cite{zitzler2001spea2} and NSGA-II \cite{deb2002fast}. The aim of the experiment is to analyze the possibility of choosing a single effective selection strategy for different model design tasks (see \hyperref[hypsel]{Hypothesis 2}).
\subsubsection*{Experiment 3} Comparison of the parameter-free GPComp@Free approach against different multi-objective EAs (based on the simple steady-state algorithm and a partial algorithm with a fixed maximum depth of the model structure graph). The aim of the experiment is to verify the effectiveness of the proposed approach against existing multi-objective schemes (\hyperref[hypself]{see Hypothesis 3}).

Every experiment consists of several stages: the evaluation of each version of the evolutionary algorithm (repeated ten times to obtain a stable result); evaluation of the prediction quality (with test sample) and computational time (measured during the evaluations and verified using an empirical performance model \cite{kalyuzhnaya2021towards}); analysis of Pareto frontiers and hypervolume values. The solid area around line on the graph shows the deviation in  quality metric for each of the proposed algorithms. For the non-parameter-free algorithms, the population size was set to 20 and the maximum number of generations was set to 30.

\subsection{Experimental results}

\subsubsection*{Experiment 1}

The first experiment is aimed to compare the multi-objective (MO) approach to the single-objective (SO) competitors. From the analysis of the quality metrics for each of the algorithms presented in Fig.~\ref{fig_algorithm_compare}, we can conclude that starting with the $10^{\text{th}}$ generation the MO algorithm consistently outperforms the SO approach and is slightly inferior to the SO approach with a penalty function. However, later in the runs, with the growth of the number of generations, the MO approach surpasses all others. An important fact that is worth mentioning is that the results of the MO algorithm are much more robust and stable in time compared to the alternative approaches. Summing up the above, it can be argued that with a sufficient number of generations, a MO algorithm is more preferable. Partial results for the 228\_elusage data set are presented in Fig.~\ref{fig_algorithm_compare}.

\begin{figure}[t]
\centerline{\includegraphics[width=9cm]{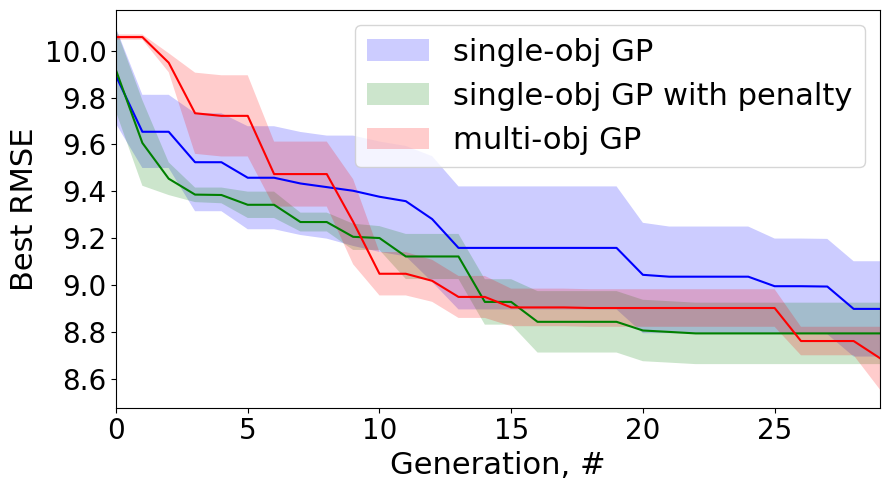}}
\caption{Comparison of RMSE objective-function value dynamics during the optimization for different algorithms: single and multi-objective (from Experiment 1). The results are obtained with the 228\_elusage data set.}
\label{fig_algorithm_compare}
\end{figure}

It can be seen that the multi-objective approach allows achieving a slightly better quality of the prediction. The values of the graph size $G_{s}$ and graph depth $G_{d}$ from Table~\ref{tab_algorithms_comparasion} are also lower for the multi-objective case that confirms its practical applicability and allows us to continue the set of experiments and confirms  \hyperref[hypmo]{Hypothesis 1}.

\subsubsection*{Experiment 2}
In the next experiment, we compared two multi-objective selection strategies. As can be seen in Figure~\ref{fig_algorithm_compare_exp2}, the value of the first objective function (in this case, ROC AUC) for the algorithm with the SPEA2-based selection is higher for each data set. It should also be noted that based on the results of ten runs for each of the algorithms, it can be concluded that SPEA2-based selection is more robust and reliable. The analysis of the number of points in Pareto frontiers (described in Table~\ref{tab_algorithms_comparasion}) confirms that the SPEA2-based selection is more suitable for further experiments since the Pareto frontiers obtained for this variant of the algorithm are more diverse and contain more points, which confirms the \hyperref[hypsel]{Hypothesis 2}.

\begin{figure}[t]
\centerline{\includegraphics[width=9cm]{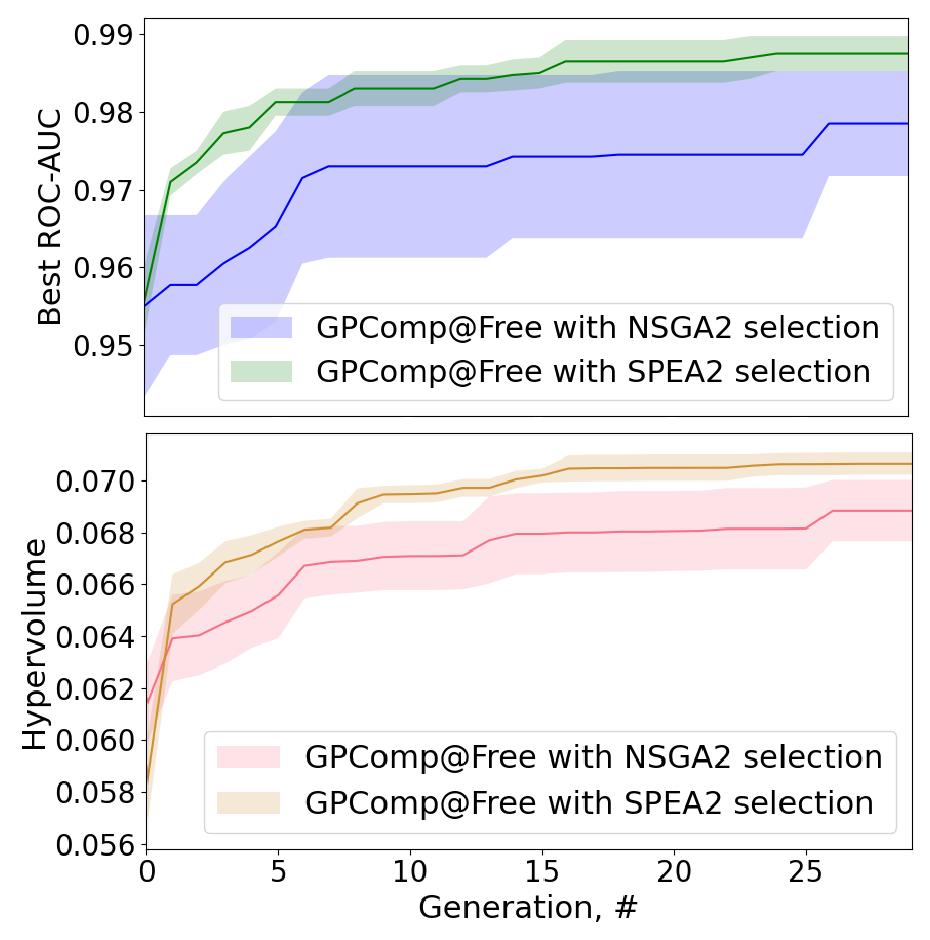}}
\caption{Comparison of objective function values convergence (top) and hypervolume values convergence (bottom) during the optimization for different selection operators (from Experiment~2). The results are obtained with the Dis data set.}
\label{fig_algorithm_compare_exp2}
\end{figure}

\subsubsection*{Experiment 3}

The final experiment was devoted to the comparison of different multi-objective approaches. The main idea of this experiment is to find out whether it is possible to get higher values of the objectives using self-configuration of the evolutionary algorithm hyperparameters during optimization. As an alternative to this method, an approach was used in which the values of hyperparameters were set before the optimization began. The results presented in Fig.~\ref{fig_config_compare} for the Hill Valley data set indicate that starting from a certain generation (in our case, from the fifth), the approach using self-configured hyperparameters of the evolutionary algorithm begins to consistently outperform the alternative approach. It should also be noted that as the number of generations increases, the self-configured algorithm provides more robust and stable solutions. The results of the experiment for all data sets that are shown in the Table~\ref{tab_algorithms_comparasion} confirm \hyperref[hypsel]{Hypothesis 3}.

\begin{figure}[t]
\centerline{\includegraphics[width=9cm]{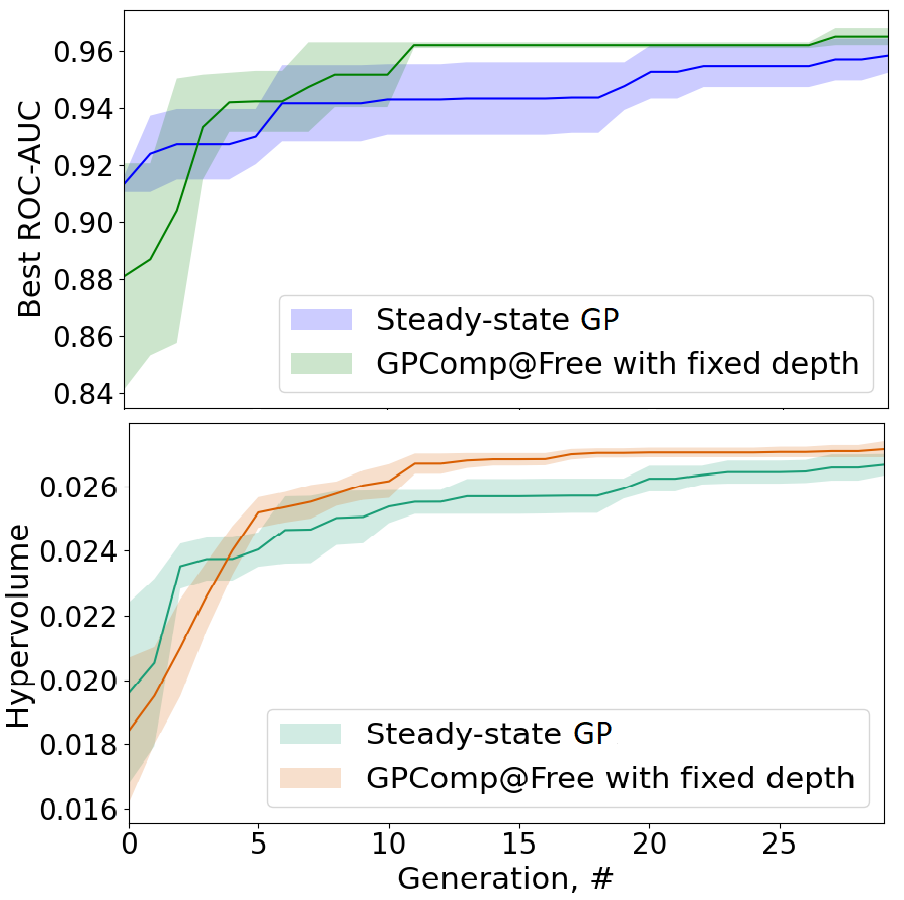}}
\caption{Comparison of objective function and hypervolume convergence during the optimization of the different types of multi-objective approaches (from Experiment 3). The two variants of the steady-state GP and two variants of the GPComp@Free are presented. The results are obtained with the Hill Valley data set.}
\label{fig_config_compare}
\end{figure}

Additional comparison of best Pareto frontiers for each approach in the experiment with the Dis data set and detailed representation of certain models is presented in Figure~\ref{fig_pareto_scheme}. It can be seen that all points in the frontier for GPComp@Free are dominating the points for the steady-state GP algorithm, which confirms its efficiency.

\begin{figure}[ht!]
\centerline{\includegraphics[width=9cm]{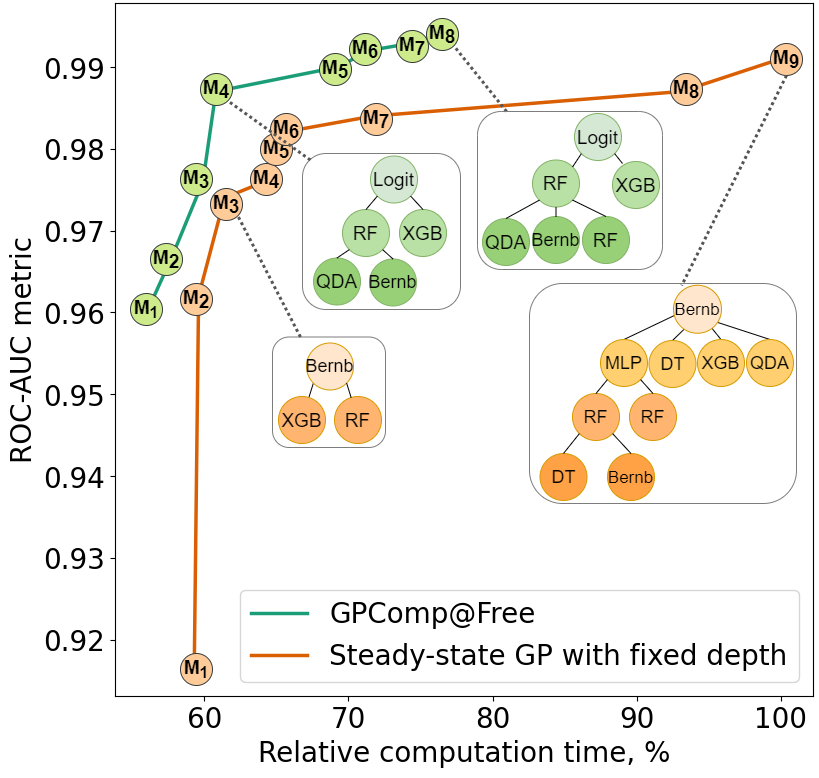}}
\caption{Pareto frontiers for multi-objective optimization with different algorithms for Dis data set. The internal structure of the obtained composite models is presented as graphs with names of the used ML models in nodes.}
\label{fig_pareto_scheme}
\end{figure}

\subsubsection*{Detailed results of experiments}

The summary of the results for all experiments and data sets is presented in Table ~\ref{tab_algorithms_comparasion}. It also includes the comparison of GPComp@Free with state-of-the-art AutoML approach TPOT (also based on GP concept) and single-model baseline Random Forest with default parameters which demonstrates its greatest efficiency. Best algorithm for each experiment and its solution characteristics highlighted with bold.

\begin{table}
\centering
\caption{Results of the experiments for different versions of the single-objective algorithm and the multi-objective algorithm. The algorithms with the complexity penalty term (PT) and the fixed depth heuristic (FD) were involved in the comparison. ROC AUC, root mean squared error (RMSE) and hypervolume (${F}_{hyp}$) were used to evaluate the solutions. $G_s$ and $G_d$ represent the size and the depth of the model structure graph. The classification and regression benchmarks from PMLB repository were used as test cases. For all experiments, the deviation in the obtained quality metric was less then 5 percent.}
\label{tab_algorithms_comparasion}
\begin{tabular}{|c|c|c|c|c|c|} 
\hline
\begin{tabular}[c]{@{}c@{}}Data\\set\end{tabular}                          & Option                           & \begin{tabular}[c]{@{}c@{}}Quality\\(ROC\\AUC\\or\\RMSE)\end{tabular} & \begin{tabular}[c]{@{}c@{}}$G_{s}$\\and\\$G_{d}$\\for\\best\end{tabular} & ${F}_{hyp}$                & $N_{f}$      \\ 
\hline
\multicolumn{6}{|c|}{Exp. \#1 - Comparison of the single-obj and multi-obj approaches}                                                                                                                                                                                                                       \\ 
\hline
\multirow{3}{*}{Churn}                                                     & \textbf{Single-objective}        & \textbf{0.932}                                                        & \textbf{15;3}                                                            & \textbf{0.142}             & \textbf{20}  \\ 
\cline{2-6}
                                                                           & Single-objective PT              & 0.921                                                                 & 6;3                                                                      & 0.139                      & 20           \\ 
\cline{2-6}
                                                                           & Multi-objective                  & 0.924                                                                 & 7;3                                                                      & 0.14                       & 5            \\ 
\hline
\multirow{3}{*}{Dis}                                                       & Single-objective                 & 0.928                                                                 & 13;3                                                                     & 0.067                      & 4            \\ 
\cline{2-6}
                                                                           & Single-objective PT              & 0.921                                                                 & 6;3                                                                      & 0.067                      & 5            \\ 
\cline{2-6}
                                                                           & \textbf{Multi-objective}         & \textbf{0.958}                                                        & \textbf{7;3}                                                             & \textbf{0.069}             & \textbf{10}  \\ 
\hline
\multirow{3}{*}{\begin{tabular}[c]{@{}c@{}}Hill\\Valley\end{tabular}}      & Single-objective                 & 0.995                                                                 & 12;3                                                                     & 0.032                      & 3            \\ 
\cline{2-6}
                                                                           & Single-objective PT              & 0.921                                                                 & 5;2                                                                      & 0.030                      & 5            \\ 
\cline{2-6}
                                                                           & \textbf{Multi-objective}         & \textbf{0.999}                                                        & \textbf{4;2}                                                             & \textbf{0.033}             & \textbf{6}   \\ 
\hline
\multirow{3}{*}{\begin{tabular}[c]{@{}c@{}}Elusage \\(regr.)\end{tabular}} & Single-objective                 & 8.477                                                                 & 10;4                                                                     & 0.235                      & 20           \\ 
\cline{2-6}
                                                                           & Single-objective PT              & 8.367                                                                 & 12;4                                                                     & 0.237                      & 20           \\ 
\cline{2-6}
                                                                           & \textbf{Multi-objective}         & \textbf{8.364}                                                        & \textbf{8;3}                                                             & \textbf{0.240}             & \textbf{4}   \\ 
\hline
\multicolumn{6}{|c|}{Exp. \#2 - comparison of the multi-obj selections types}                                                                                                                                                                                                                                \\ 
\hline
\multirow{2}{*}{Churn}                                                     & NSGA selection                   & 0.924                                                                 & 5;3                                                                      & 0.140                      & 4            \\ 
\cline{2-6}
                                                                           & \textbf{SPEA2 selection}         & \textbf{0.929}                                                        & \textbf{5;3}                                                             & \textbf{0.140}             & \textbf{10}  \\ 
\hline
\multirow{2}{*}{Dis}                                                       & NSGA selection                   & 0.994                                                                 & 6;3                                                                      & 0.067                      & 5            \\ 
\cline{2-6}
                                                                           & \textbf{SPEA2 selection}         & \textbf{0.995}                                                        & \textbf{6;3}                                                             & \textbf{0.069}             & \textbf{5}   \\ 
\hline
\multirow{2}{*}{\begin{tabular}[c]{@{}c@{}}Hill\\Valley\end{tabular}}      & NSGA selection                   & 0.982                                                                 & 7;3                                                                      & 0.032                      & 4            \\ 
\cline{2-6}
                                                                           & \textbf{SPEA2 selection}         & \textbf{0.986}                                                        & \textbf{6;3}                                                             & \textbf{0.033}             & \textbf{6}   \\ 
\hline
\multirow{2}{*}{\begin{tabular}[c]{@{}c@{}}Elusage \\(regr.)\end{tabular}} & NSGA selection                   & 8.413                                                                 & 13;3                                                                     & 0.236                      & 4            \\ 
\cline{2-6}
                                                                           & \textbf{SPEA2 selection}         & \textbf{8.314}                                                        & \textbf{6;3 }                                                            & \textbf{0.240 }            & \textbf{5 }  \\ 
\hline
\multicolumn{6}{|c|}{Exp. \#3 - Comparison of multi-obj algorithms}                                                                                                                                                                                                                                          \\ 
\hline
\multirow{4}{*}{Churn}                                                     & \textbf{Paremeter-free with FD } & \textbf{0.929}                                                        & \textbf{8;3 }                                                            & \textbf{0.435}             & \textbf{10}  \\ 
\cline{2-6}
                                                                           & GPComp@Free                      & 0.928                                                                 & 6;3                                                                      & 0.430                      & 8            \\ 
\cline{2-6}
                                                                           & Steady-state with FD             & 0.923                                                                 & 7;3                                                                      & 0.427                      & 7            \\ 
\cline{2-6}
                                                                           & Steady-state                     & 0.928                                                                 & 6;3                                                                      & 0.427                      & 8            \\ 
\hline
\multirow{4}{*}{Dis}                                                       & Parameter-free with FD           & 0.992                                                                 & 8;3                                                                      & 0.761                      & 8            \\ 
\cline{2-6}
                                                                           & \textbf{GPComp@Free}             & \textbf{0.994}                                                        & \textbf{6;3}                                                             & \textbf{0.771}             & \textbf{7}   \\ 
\cline{2-6}
                                                                           & Steady-state with FD             & 0.991                                                                 & 6;3                                                                      & 0.756                      & 8            \\ 
\cline{2-6}
                                                                           & Steady-state                     & 0.99                                                                  & 5;2                                                                      & 0.758                      & 6            \\ 
\hline
\multirow{4}{*}{\begin{tabular}[c]{@{}c@{}}Hill\\Valley\end{tabular}}      & Parameter-free with FD           & 0.972                                                                 & 8;4                                                                      & 0.032                      & 4            \\ 
\cline{2-6}
                                                                           & \textbf{GPComp@Free}             & \textbf{0.974}                                                        & \textbf{5;3}                                                             & \textbf{0.032}             & \textbf{5}   \\ 
\cline{2-6}
                                                                           & Steady-state with FD             & 0.960                                                                 & 6;3                                                                      & 0.030                      & 4            \\ 
\cline{2-6}
                                                                           & Steady-state                     & 0.973                                                                 & 5;3                                                                      & 0.031                      & 5            \\ 
\hline
\multirow{4}{*}{\begin{tabular}[c]{@{}c@{}}Elusage \\(regr.)\end{tabular}} & \textbf{Parameter-free with FD}  & \textbf{8.583}                                                        & \textbf{7;3 }                                                            & \textbf{0.223}             & \textbf{5 }  \\ 
\cline{2-6}
                                                                           & GPComp@Free                      & 8.762                                                                 & 10;3                                                                     & 0.219                      & 7            \\ 
\cline{2-6}
                                                                           & Steady-state with FD             & 8.825                                                                 & 7;3                                                                      & 0.198                      & 6            \\ 
\cline{2-6}
                                                                           & Steady-state                     & 9.211                                                                 & 6;3                                                                      & \multicolumn{1}{l|}{0.137} & 4            \\ 
\hline
\multicolumn{6}{|c|}{Comparasion with state-of-art and baseline}                                                                                                                                                                                                                                             \\ 
\hline
\multirow{3}{*}{Churn}                                                     & TPOT                             & 0.908                                                                 & 2;1                                                                      & -                          & -            \\ 
\cline{2-6}
                                                                           & Random forest                    & 0.887                                                                 & 1;1                                                                      & -                          & -            \\ 
\cline{2-6}
                                                                           & \textbf{GPComp@Free}             & \textbf{0.924}                                                        & \textbf{7;3}                                                             & \textbf{0.14}              & \textbf{5}   \\ 
\hline
\multirow{3}{*}{Dis}                                                       & TPOT                             & 0.932                                                                 & 3;1                                                                      & -                          & -            \\ 
\cline{2-6}
                                                                           & Random forest                    & 0.918                                                                 & 1;1                                                                      & -                          & -            \\ 
\cline{2-6}
                                                                           & \textbf{GPComp@Free}             & \textbf{0.994}                                                        & \textbf{6;3}                                                             & \textbf{0.771}             & \textbf{7}   \\ 
\hline
\multirow{3}{*}{\begin{tabular}[c]{@{}c@{}}Hill\\Valley\end{tabular}}      & TPOT                             & 0.999                                                                 & 3;1                                                                      & -                          & -            \\ 
\cline{2-6}
                                                                           & Random forest                    & 0.638                                                                 & 1;1                                                                      & -                          & -            \\ 
\cline{2-6}
                                                                           & \textbf{GPComp@Free}             & \textbf{0.999}                                                        & \textbf{4;2}                                                             & \textbf{0.033}             & \textbf{6}   \\ 
\hline
\multirow{3}{*}{\begin{tabular}[c]{@{}c@{}}Elusage\\(regr.)\end{tabular}}  & TPOT                             & 11.682                                                                & 3;1                                                                      & -                          & -            \\ 
\cline{2-6}
                                                                           & Random forest                    & 13.417                                                                & 1;1                                                                      & -                          & -            \\ 
\cline{2-6}
                                                                           & \textbf{GPComp@Free}             & \textbf{8.364}                                                        & \textbf{8;3}                                                             & \textbf{0.240}             & \textbf{4}   \\
\hline
\end{tabular}
\end{table}

As can be seen in this table, the results obtained during the experiments demonstrate the advantage of multi-objective optimization algorithms over other used approaches. The only exception is a single case from the first experiment, where the maximum value of the quality metric was obtained using a single-objective algorithm. However, it should be noted that the difference in the obtained values of the quality metric for the single-objective and multi-objective algorithms is less than 0.01. At the same time, a much more simple model structure was designed in the case of a multi-objective approach. In the second experiment, the SPEA2 selection showed higher results for each of the three selected data sets. Thus, this algorithm was chosen for the next experiments. Results obtained during the third experiment confirm that the parameter-free optimization approach GPComp@Free allows achieving better results compared to the other approaches. The version of the algorithm with the non-fixed depth of the final composite model did not decrease the values of the quality metric and hypervolume. It allows us to obtain more simple composite models (from both structural and computational points of view) without loss of prediction quality.

\section{Conclusion and discussions}
\label{sec_disc}

There are a lot of different decisions that should be made to build an effective, robust, and flexible algorithm for the design of the composite models or modeling pipelines. In the paper, we propose the self-tuning approach that can be effectively used without additional meta-optimization, which is critically important for the application in AutoML-related solutions since it allows achieving a better degree of automation. 

The results of the experiments confirm that the proposed evolutionary approach GPComp@Free allows achieving a stable advantage against different single-objective and multi-objective implementations. It can be seen that the composite models found with a parameter-free approach are more effective and less complex according to the various benchmarks.

The proposed approach is not limited to the problem of the composite models' design or specific AutoML framework and can be applied as a part of various automated modeling solutions (e.g. NAS, equation discovery, etc). 

\section*{Code and data availability}

The implemented approach is available as a part of the open-source FEDOT framework\footnote{\url{https://github.com/nccr-itmo/FEDOT}}. Data and scripts that were used to conduct the experiments in the paper can also be obtained from the open repository\footnote{\url{https://github.com/ITMO-NSS-team/FEDOT-benchmarks}}.

\section*{Acknowledgment}

This research is financially supported by The Russian Scientific Foundation, Agreement \#19-11-00326.


\bibliographystyle{IEEEtran}
\bibliography{ref.bib}

\end{document}